\documentclass[letterpaper, 10 pt, conference]{ieeeconf}  

\IEEEoverridecommandlockouts                              

\usepackage{graphics} 
\usepackage{epsfig} 
\usepackage{mathptmx} 
\usepackage{times} 
\usepackage{amsmath} 
\usepackage{amssymb}  
\usepackage{dsfont}
\usepackage{makecell}

\usepackage{hyperref}

\usepackage{eurosym}
\usepackage{cite}
\usepackage{textcomp}
\usepackage{xspace}
\usepackage{xcolor} 
\usepackage{censor}
\usepackage{multirow}
\usepackage{booktabs} 
\usepackage{array}
\usepackage{placeins}
\usepackage{flushend}
\usepackage{dblfloatfix}

\usepackage[ruled,vlined,linesnumbered]{algorithm2e}
\usepackage[dvipsnames]{xcolor}

\SetCommentSty{mycommfont}

\usepackage{blindtext}

\usepackage[font=small,labelfont=bf]{caption}

\title{\LARGE \bf
Robust Surgical Robotic Instrument Tracking via \\
Sequential Multi-Cue Fusion and Sim-to-Real Self-Training
}

\author{
Hanyang Hu$^{1}$, Zekai Liang$^{1}$, Florian Richter$^{1}$, Michael C. Yip$^{1}$, \textit{Senior Member, IEEE}
\thanks{$^{1}$Department of Electrical and Computer Engineering, University of California San Diego, La Jolla, CA 92093 USA.
        {\tt\small \{hanyang-hu, z9liang, frichter, yip\}@ucsd.edu}}%
}

\DeclareMathAlphabet{\mathcal}{OMS}{cmsy}{m}{n}

\begin{document}

\maketitle
\thispagestyle{empty}
\pagestyle{empty}

\begin{abstract}
Efficient and robust tracking of surgical robotic instruments is important for robot-assisted minimally invasive surgery, yet remains challenging due to the complexity of surgical scenes and the unconventional geometry of surgical instruments.
Keypoint-based approaches are efficient, but their performance depends on reliable feature detection. Improving these detectors with real-world supervision is difficult because accurate real-world annotations are costly to obtain at scale. To address this limitation, we introduce a tracker-guided self-training framework that adapts a model pretrained on synthetic images to unlabeled real-world videos. Given measured robot joint states, an uncertainty-aware EKF recursively corrects the instrument pose and the observable joint angles by comparing projected model features with detected keypoints, shaft boundaries, and mask-derived cues. An RTS smoother subsequently refines the resulting trajectory, which is projected into pseudo-labels for fine-tuning the feature detector without laborious pose annotations. Experiments on real-world videos demonstrate consistent improvements from self-training across all evaluated keypoint metrics, and the resulting model outperforms prior approaches in both accuracy and runtime. \textit{The code and data will be released upon publication.}
\end{abstract}

\section{INTRODUCTION}

Robot-assisted minimally invasive surgery (RMIS) has gained increasing attention for improving surgical safety and efficiency while reducing surgeon fatigue. Accurate, low-latency estimation of articulated instrument pose relative to the endoscopic camera is a fundamental perception capability in RMIS. Reliable pose feedback supports downstream tasks including visual servoing~\cite{molnar2020visual}, augmented-reality guidance~\cite{kalia2021preclinical}, and autonomous tissue manipulation~\cite{li2020super}.


Modern approaches to markerless surgical robotic instrument pose estimation generally follow one of two paradigms: keypoint detection with PnP-based pose reconstruction~\cite{richter2022robotictooltracking, lu2022pose, dambrosia2024robustsurgical, liang2025efficientsurgicalroboticinstrument}, or instrument segmentation followed by iterative render-and-match optimization~\cite{liang2025differentiablerenderingbasedposeestimation, chen2025surgipose, hu2026realtimerenderingbasedsurgicalinstrument, wu2026surgcalib}. Given the computational complexity of rendering-based optimization methods, keypoint-based pose estimation methods are generally better suited to real-time robot control due to their fast inference speed. However, their accuracy depends heavily on reliable visual feature detection, which is prone to suboptimal endoscopic imaging conditions.

Compared with general robotics, the scarcity of accurately annotated real-world data makes improving feature detection robustness for surgical robots particularly challenging.
Prior work has explored synthetic data generation with photorealistic rendering to alleviate the scarcity of annotated real-world data~\cite{lu2022pose, liang2025efficientsurgicalroboticinstrument}. Although these approaches have demonstrated promising feasibility and performance gains, a substantial sim-to-real gap remains. Synthetic data alone is often insufficient to achieve the robustness required for reliable deployment in real-world applications.

\begin{figure}[!t]
    \centering
    \setlength{\tabcolsep}{1pt}
    \renewcommand{\arraystretch}{0}
    \begin{tabular}{@{}cc@{}}
        \textbf{\small Synthetic Pretrained} &
        \textbf{\small Ours (Self-Trained)} \\         \includegraphics[width=0.49\linewidth]{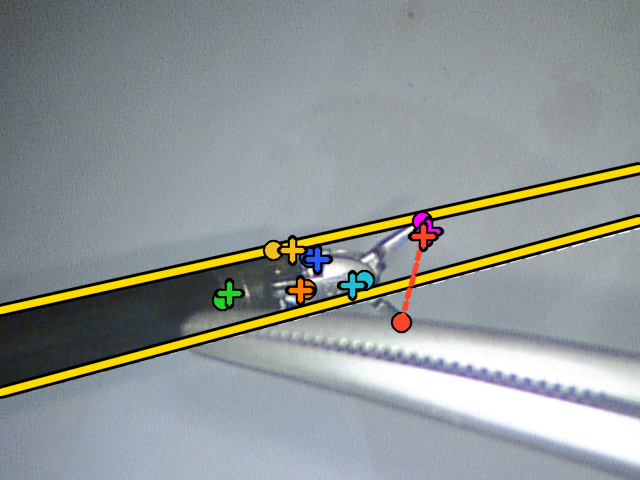} &
        \includegraphics[width=0.49\linewidth]{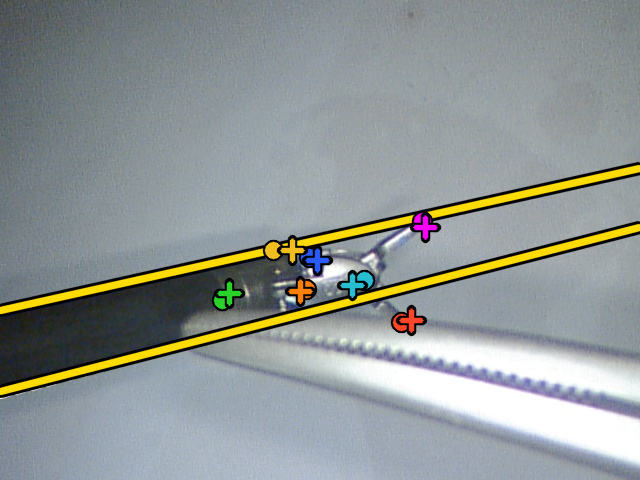} \\[1pt]
        \includegraphics[width=0.49\linewidth]{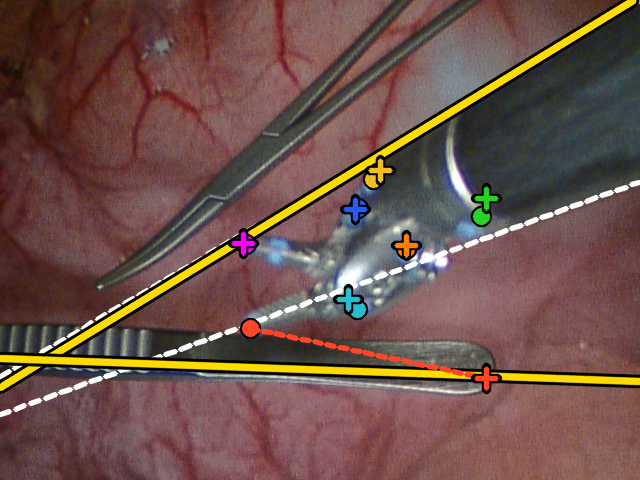} &
        \includegraphics[width=0.49\linewidth]{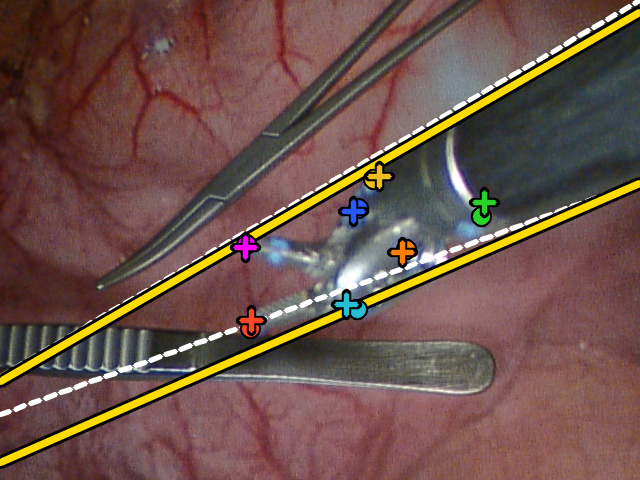} \\[3pt]

    \end{tabular}
    \caption{\textbf{Self-training improves sim-to-real feature detection.} Pretrained and self-trained models are compared on unseen real-world images with diverse orientations, distractions, and occlusions. Circles and dashed white lines denote manually labeled ground truth; plus signs and solid yellow lines denote model predictions. A tracker-guided self-training framework enables a synthetically pretrained model to bridge the gap to challenging, real-world scenes without tedious human labeling.}
    \label{fig:cover}
    \vspace{-0.18in}
\end{figure}

To address the scarcity of annotated real-world data in surgical robotics, we present a self-training framework for surgical instrument pose reconstruction that enables
sim-to-real transfer
using unlabeled real-world images. During adaptation, a distribution-aware Extended Kalman Filter (EKF)~\cite{kalman1961new}, followed by a Rauch-Tung-Striebel (RTS) smoother~\cite{rauch1965maximum}, aggregates observations across unlabeled real-world trajectories to obtain robust and temporally consistent pose estimates. These estimates are then projected back into the image space to generate pseudo-labels, which provide supervision for adapting the model to the target domain without requiring ground-truth annotations (Fig.~\ref{fig:cover}).


\begin{figure*}[t!]
\centering
\includegraphics[width=\textwidth]{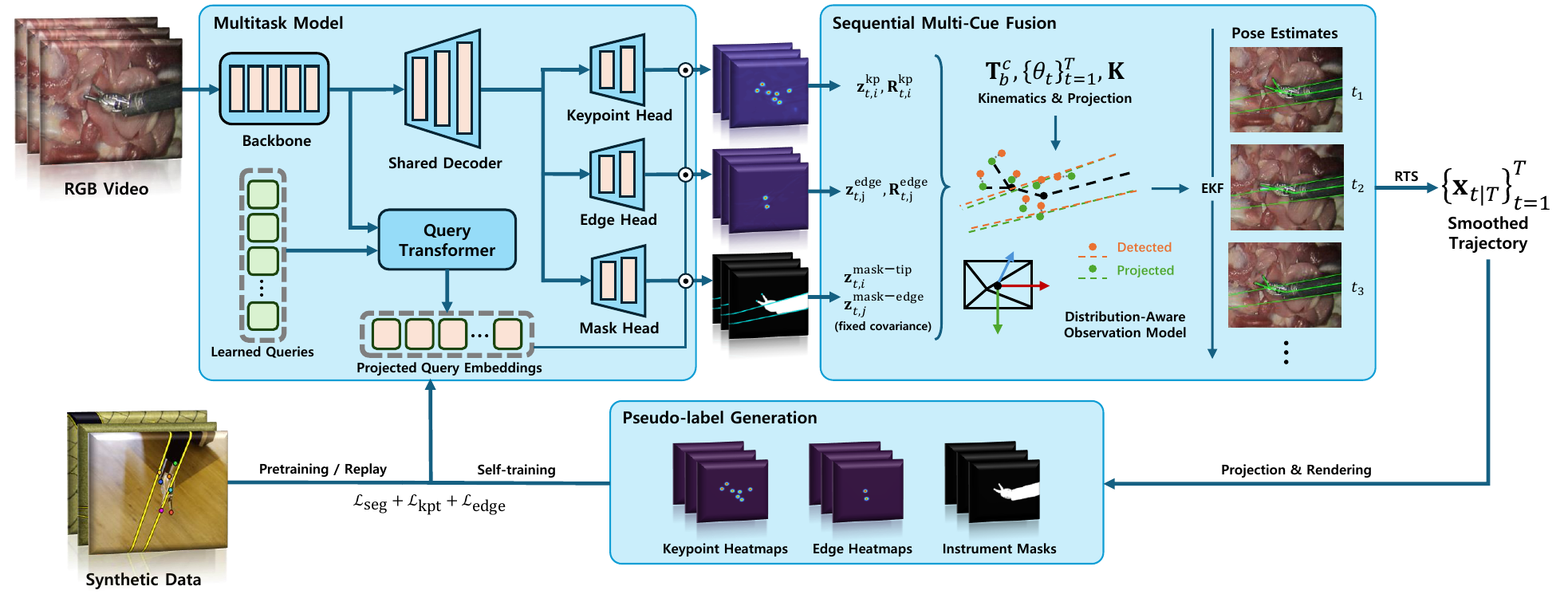}
\caption{\textbf{Overview of the proposed self-training framework.} A query-based \textbf{multitask model} predicts keypoint heatmaps, shaft edge heatmaps, and instrument masks from RGB video, which are decoded into geometric observations with associated uncertainty. \textbf{Sequential multi-cue fusion} combines these observations with robot kinematics to obtain smoothed pose trajectories, using an initial base-to-camera transform $\mathbf{T}_b^c$ estimated by differentiable rendering~\cite{liang2025differentiablerenderingbasedposeestimation}. An EKF propagates the instrument state between frames and updates it using the current frame's observations, while a backward RTS smoother refines these pose estimates to produce a smoothed pose trajectory. \textbf{Pseudo-label generation} projects and renders the smoothed instrument configurations into feature labels for sim-to-real adaptation.}
\label{fig:pipeline}
 \vspace{-0.14in}
\end{figure*}
The main contributions of this work are as follows:
\begin{enumerate}
    \item A self-training framework that leverages unlabeled real-world images to improve the accuracy and robustness of surgical robotic instrument feature detection.
    \item An efficient, distribution-aware EKF that fuses complementary keypoints, shaft edges, and mask-derived observations for robust instrument pose tracking.
    \item A pose-to-feature pseudo-labeling strategy that converts smoothed trajectories into feature annotations with consistency-based sample weighting.
\end{enumerate}



\section{Related Work}

Accurate tracking of articulated surgical instruments is challenging due to partial visibility, cable-driven kinematic errors, and limited endoscopic image quality. Earlier approaches relied on external sensing or fiducial markers~\cite{ozguner2020camera, hwang2020applying, hwang2020efficiently, wu2021closed}, or explicit modeling of transmission errors~\cite{miyasaka2015measurement} and calibrating the remote center of motion (RCM)~\cite{zhong2020hand}, whereas recent markerless methods estimate instrument pose from visual observations using either feature-based or rendering-based approaches. Among feature-based methods, \cite{richter2022robotictooltracking} model camera-to-robot calibration and kinematic errors as a lumped error transform, which they track using visual landmarks and shaft boundary observations. \cite{dambrosia2024robustsurgical} extend this framework with learned shaft edge detection and pixel-based probabilistic observation models, while \cite{liang2025efficientsurgicalroboticinstrument} introduce unified keypoint and edge detection together with an efficient geometry-based pose solver. EKFs have also been used to fuse visual observations with robot kinematics for instrument tracking and online hand-eye calibration~\cite{ye2016real,moccia2020vision, shinde2024surestep, cui2026ontheflyhandeyecalibrationda}. On the other hand, rendering-based methods exploit dense instrument shape information by comparing rendered silhouettes with instrument segmentation through iterative optimization~\cite{liang2025differentiablerenderingbasedposeestimation, chen2025surgipose, hu2026realtimerenderingbasedsurgicalinstrument, wu2026surgcalib}. \cite{liang2025differentiablerenderingbasedposeestimation} optimize instrument pose through differentiable rendering and demonstrate improved robustness over PnP-based methods, while \cite{hu2026realtimerenderingbasedsurgicalinstrument} achieve real-time tracking using evolutionary optimization and batch rendering. However, rendering-based methods remain computationally demanding and are sensitive to incomplete or noisy segmentation caused by occlusion and complex surgical scenes.

Outside surgical robotics, self-supervised learning and self-training have been explored to adapt pose estimation models using unlabeled real-world data. CtRNet~\cite{lu2023markerlesscameratorobotposeestimation,lu2024ctrnetxcameratorobotposeestimation} uses differentiable pose estimation and rendering for self-supervision to improve segmentation and keypoint detection on unlabeled real-world images. More closely related to tracker-supported self-training, \cite{lu2022slamsupported} fuse object pose predictions with robot odometry in a pose graph to generate temporally consistent pseudo-labels. \cite{park2024onlinesupervised} use an adaptive unscented Kalman filter to refine spacecraft pose estimates for online self-training.
These works demonstrate how geometric and temporal consistency can provide supervision for pose learning without laborious data annotation.
However, these approaches have not been adapted to RMIS, where partial visibility of the instrument, suboptimal endoscopic imaging conditions, and complex tissue backgrounds make both feature detection and pose tracking substantially less reliable.

\section{Methodology}

An overview of the proposed framework is illustrated in Fig.~\ref{fig:pipeline}. 
The framework comprises two main stages: synthetic pretraining of a multitask detector, followed by tracker-guided self-training on unlabeled real-world videos. 
During inference, the adapted feature detector provides observations to the forward EKF for online pose estimation.

\subsection{Synthetic Pretraining}

\subsubsection{Data Generation}

\begin{figure}[!t]
    \centering
    \setlength{\tabcolsep}{1pt}
    \renewcommand{\arraystretch}{0}
    \begin{tabular}{@{}cc@{}}
        \includegraphics[width=0.49\linewidth]{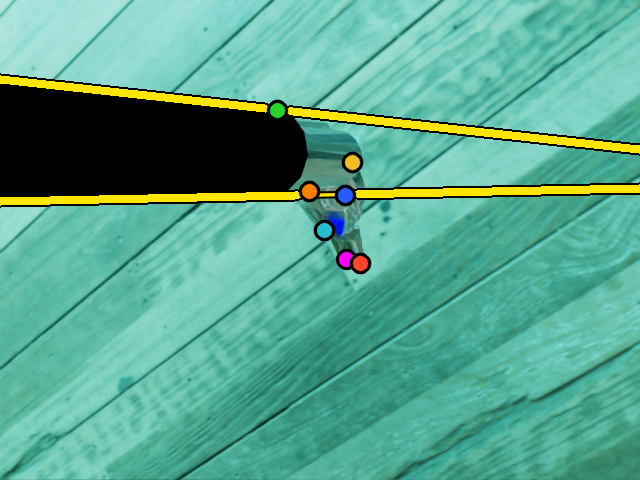} &
        \includegraphics[width=0.49\linewidth]{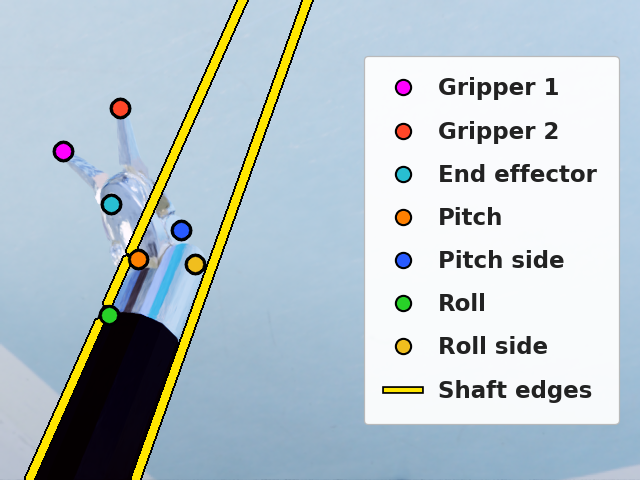} \\[1pt]
        \includegraphics[width=0.49\linewidth]{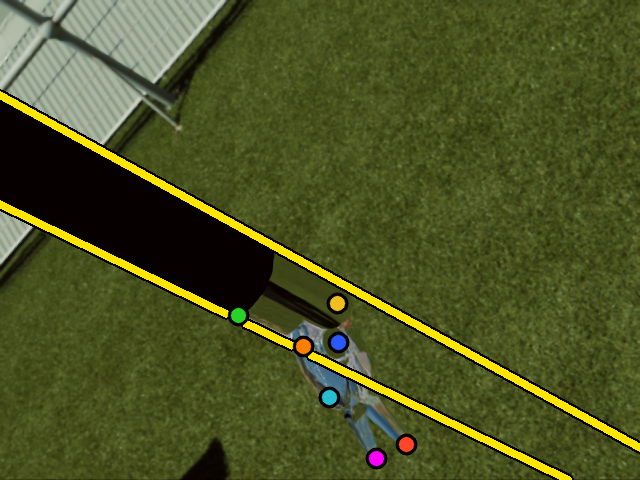} &
        \includegraphics[width=0.49\linewidth]{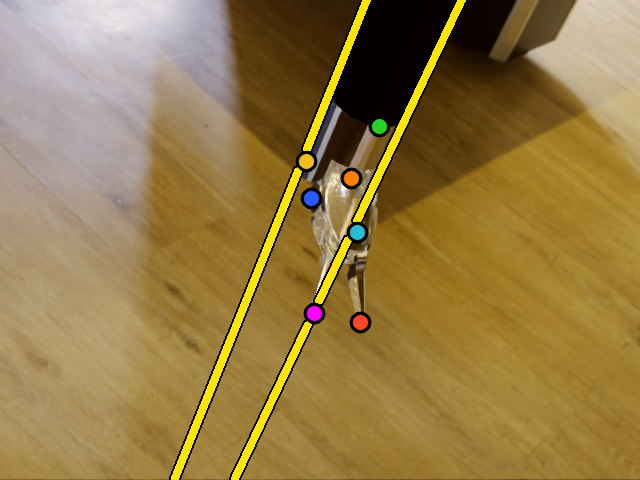}
    \end{tabular}
    \caption{\textbf{Synthetic data generation.} Isaac Sim generates labeled images with ground-truth keypoints and shaft edge annotations.}
    \label{fig:isaac_data_examples}
    \vspace{-0.5in}
\end{figure}

NVIDIA Isaac Sim is used to render synthetic RGB images and generate the corresponding ground-truth instrument masks, semantic keypoints, and shaft edge annotations for pretraining (Fig.~\ref{fig:isaac_data_examples}). The keypoint definitions are adapted from the SurgPose dataset~\cite{wu2025surgposedatasetarticulatedrobotic}, consisting of two gripper tips and five landmarks on the visible surface of the instrument. 
These visually distinctive landmarks jointly constrain the instrument articulation and shaft orientation. Similar to~\cite{liang2025efficientsurgicalroboticinstrument}, the two projected edges of the cylindrical shaft serve as complementary features for pose reconstruction.

To cover diverse viewpoints and instrument articulations in the synthetic training dataset, the instrument pose and joint configuration are sampled over the pose hypothesis space defined in~\cite{liang2025differentiablerenderingbasedposeestimation}. The data generation pipeline further randomizes the illumination and background appearance and applies heavy photometric augmentation to improve robustness to the sim-to-real gap.

\subsubsection{Network Architecture}
\label{sec:network}

The detector uses a pretrained DINOv3-B backbone~\cite{dinov3} and a MaskFormer-inspired~\cite{cheng2021perpixelclassificationneedsemantic} query-based multitask prediction module, which comprises a shared image-space decoder, a query Transformer, and three task-specific branches that predict keypoint heatmaps, edge heatmaps, and instrument segmentation.

Given an RGB image of resolution $224\times224$, DINOv3 produces a $14\times14$ grid of patch token embeddings. Starting from the final Transformer block (block 12), the image-space decoder progressively upsamples the features and fuses lateral projections from blocks 9, 6, and 3, yielding a shared representation $\mathbf{F}_{\mathrm{shared}}\in\mathbb{R}^{112\times112\times128}$.
Separate mask and keypoint adapters increase the spatial resolution of $\mathbf{F}_{\mathrm{shared}}$ through bilinear upsampling and convolution to produce $\mathbf{F}_{\mathrm{mask}},\mathbf{F}_{\mathrm{kp}} \in\mathbb{R}^{224\times224\times128}$. The edge adapter instead downsamples $\mathbf{F}_{\mathrm{shared}}$ and maps it through a Deep Hough Transform (DHT)~\cite{zhao2021deep} layer, followed by convolution and upsampling, to obtain $\mathbf{F}_{\mathrm{edge}}\in\mathbb{R}^{224\times224\times16}$ in Hough space.

Following MaskFormer's query-based design~\cite{cheng2021perpixelclassificationneedsemantic}, we introduce ten learnable query tokens: one mask query, seven keypoint queries, and two edge queries. These tokens are jointly processed by a Transformer decoder that cross-attends to the $196$ patch tokens from the final DINOv3 backbone layer.
Task-specific linear projections then map the decoded queries to the corresponding feature dimensions, yielding $\mathbf{q}_{\mathrm{mask}}$, $\{\mathbf{q}^{\mathrm{kp}}_i\}_{i=1}^{7}$, and $\{\mathbf{q}^{\mathrm{edge}}_j\}_{j=1}^{2}$.
Predictions are formed by dot products between the projected query embeddings and their corresponding dense features, with a learned bias for each query
\begin{align}
    \mathbb{S}(\mathbf{u})
    &=
    \sigma\!\left(
        \left\langle
        \mathbf{q}_{\mathrm{mask}},
        \mathbf{F}_{\mathrm{mask}}(\mathbf{u})
        \right\rangle
        + b_{\mathrm{mask}}
    \right), \\
    \mathbb{H}_{i}(\mathbf{u})
    &=
    \left\langle
    \mathbf{q}^{\mathrm{kp}}_{i},
    \mathbf{F}_{\mathrm{kp}}(\mathbf{u})
    \right\rangle
    + b^{\mathrm{kp}}_{i},
    \qquad i=1,\ldots,7, \\
    \mathbb{E}_{j}(\boldsymbol{\eta})
    &=
    \left\langle
    \mathbf{q}^{\mathrm{edge}}_{j},
    \mathbf{F}_{\mathrm{edge}}(\boldsymbol{\eta})
    \right\rangle
    + b^{\mathrm{edge}}_{j},
    \qquad j=1,2,
\end{align}
where $\mathbf{u}$ indexes image-space locations and $\boldsymbol{\eta}=[\theta,\rho]^\top$ indexes locations in the discretized orientation-offset Hough space. This produces the instrument mask $\mathbb{S}\in\mathbb{R}^{224\times224}$,
seven keypoint heatmaps $\mathbb{H}\in\mathbb{R}^{224\times224\times7}$,
and two shaft edge heatmaps $\mathbb{E}\in\mathbb{R}^{224\times224\times2}$.

\subsubsection{Training Objective}

Let $\mathbb{S}^\ast$ denote the ground-truth instrument mask. The segmentation branch is trained using binary cross-entropy and Dice losses
\begin{equation}
    \mathcal{L}_{\mathrm{seg}}
    =
    \mathcal{L}_{\mathrm{BCE}}(\mathbb{S},\mathbb{S}^\ast)
    +
    \mathcal{L}_{\mathrm{Dice}}(\mathbb{S},\mathbb{S}^\ast).
\end{equation}

The ground-truth keypoints and shaft edges are encoded as Gaussian target heatmaps $\{\mathbb{H}_i^\ast\}_{i=1}^7$ and $\{\mathbb{E}_j^\ast\}_{j=1}^2$ centered at their corresponding image-space and discretized Hough-space locations, respectively. The feature branches are trained using a foreground-weighted Adaptive Wing loss $\mathcal{L}_{\mathrm{AW}}(\cdot, \cdot)$~\cite{wang2019adaptive}, which emphasizes locations near the target features. The unordered gripper tip and shaft edge heatmap regression losses are each minimized over the two possible assignments $\mathcal{S} = \{\mathrm{id},\mathrm{swap}\}$, while the remaining keypoints retain fixed correspondences
\begin{align}
    \mathcal{L}_{\mathrm{kp}}
    &=
    \min_{\pi\in \mathcal{S}}
    \sum_{i=1}^{2}
    \mathcal{L}_{\mathrm{AW}}
    \left(
        \mathbb{H}_i,
        \mathbb{H}_{\pi(i)}^{\ast}
    \right)
    +
    \sum_{i=3}^{7}
    \mathcal{L}_{\mathrm{AW}}
    \left(
        \mathbb{H}_i,
        \mathbb{H}_i^{\ast}
    \right),
    \\
    \mathcal{L}_{\mathrm{edge}}
    &=
    \min_{\pi\in\mathcal{S}}
    \sum_{j=1}^{2}
    \mathcal{L}_{\mathrm{AW}}
    \left(
        \mathbb{E}_j,
        \mathbb{E}_{\pi(j)}^{\ast}
    \right).
\end{align}
These losses define the per-sample training objective
\begin{equation}
    \mathcal{L}_{\mathrm{train}}
    =
    \mathcal{L}_{\mathrm{seg}}
    +
    \mathcal{L}_{\mathrm{kp}}
    +
    \mathcal{L}_{\mathrm{edge}}
    \label{eqn:syn_loss}
\end{equation}
which is averaged over the samples in each training batch.

\subsection{Real-World Self-Training}


For each unlabeled real-world video, the pretrained model predicts the keypoint heatmaps $\{\mathbb{H}_{t,i}\}_{i=1}^7$, shaft edge heatmaps $\{\mathbb{E}_{t,j}\}_{j=1}^2$, and instrument mask $\mathbb{S}_t$ at each frame~$t$. The base-to-camera transform $\mathbf{T}_b^c$ is estimated from the initial mask prediction through differentiable rendering~\cite{liang2025differentiablerenderingbasedposeestimation}. An EKF 
subsequently fuses the keypoint, shaft edge, and mask-derived observations with the robot kinematics, jointly estimating a 9-DoF correction to the instrument configurations~\cite{richter2022robotictooltracking}. After the forward filtering pass, an RTS smoother refines the estimated state trajectory, and the smoothed configurations are projected into the image to generate temporally consistent pseudo-labels for self-training.

\subsubsection{State Parameterization}

The instrument pose is defined at the wrist pitch frame of the dVRK PSM~\cite{kazanzides2014open} and denoted by $\mathbf{T}_{r,t}^{c}\in\mathrm{SE}(3)$ at frame $t$. Let 
$\theta_t^{(1)},\ldots,\theta_t^{(7)}$
denote the joint readings at frame $t$. The first four joint readings determine this pose through forward kinematics, while the last three joint readings describe the wrist pitch, wrist yaw, and jaw articulations. Specifically, given an estimated base-to-camera transform $\mathbf{T}_{b}^{c}$, the nominal instrument pose is
\begin{equation}
    \widetilde{\mathbf{T}}_{r,t}^{c}
    =
    \mathbf{T}_{b}^{c}
    \mathbf{T}_{r}^{b}
    \left(\theta_t^{(1:4)}\right).
    \label{eqn:nominal_pose}
\end{equation}
The accuracy of (\ref{eqn:nominal_pose}) is sensitive to base-to-camera calibration errors and transmission errors in the first four joint readings. Following~\cite{richter2022robotictooltracking}, the combined effect of these errors is modeled as a lumped error state $\boldsymbol{\xi}_t =[\alpha_t,\beta_t,\gamma_t,\mathbf{b}_t^{\top}]^{\top} \in \mathbb{R}^6$. The corrected pose estimate is given by
\begin{equation}
    \widehat{\mathbf{T}}_{r,t}^{c}(\boldsymbol{\xi}_t)
    =
    \mathbf{T}_{b}^{c}
    \Delta\mathbf{T}(\boldsymbol{\xi}_t)
    \mathbf{T}_{r}^{b}
    \left(\theta_t^{(1:4)}\right)
    \label{eqn:estimated_pose}
\end{equation}
where $\Delta\mathbf{T}(\boldsymbol{\xi}_t)\in\mathrm{SE}(3)$ denotes the pose correction with translation $\mathbf{b}_t\in\mathbb{R}^{3}$ and rotation $\mathbf{R}_y(\gamma_t)\mathbf{R}_x(\alpha_t)\mathbf{R}_z(\beta_t)$, using the look-at parameterization proposed in~\cite{liang2025differentiablerenderingbasedposeestimation}.

In addition, errors in the three visible joint angles are separately modeled by  $\delta\boldsymbol{\theta}_t = [\delta \theta_t^{(5)},\delta \theta_t^{(6)},\delta \theta_t^{(7)}]^{\top}$ with
\begin{equation}
    \widehat{\theta}_t^{(k)}=\theta_t^{(k)}+\delta \theta_t^{(k)}, \qquad k=5,6,7.
\end{equation}
Together, the pose and joint corrections form the nine-dimensional EKF state $\mathbf{x}_t = [\boldsymbol{\xi}_t^{\top}, \delta\boldsymbol{\theta}_t^{\top}]^{\top} \in \mathbb{R}^{9}$.
Under this formulation, state evolution is modeled by a random walk~\cite{richter2022robotictooltracking}
\begin{equation}
    \mathbf{x}_{t+1}
    =
    \mathbf{x}_{t}+\mathbf{w}_{t},
    \qquad
    \mathbf{w}_{t}\sim
    \mathcal{N}(\mathbf{0},\mathbf{Q}),
    \label{eqn:motion_model}
\end{equation}
where $\mathbf{w}_{t}$ is the process noise and $\mathbf{Q} \in \mathbb{R}^{9\times9}$ is the process noise covariance.

\begin{algorithm}[t]
    \footnotesize
    \caption{Sequential Multi-Cue Fusion}
    \label{alg:multi_cue_fusion}

    \KwIn{Model predictions
    $\{\mathbb{S}_t,\mathbb{H}_t,\mathbb{E}_t\}_{t=1}^{T}$,
    joint readings $\{\boldsymbol{\theta}_t\}_{t=1}^{T}$,
    base-to-camera transform $\mathbf{T}_b^c$,
    camera intrinsics $\mathbf{K}$,
    initial belief
    $(\mathbf{x}_{0|0},\boldsymbol{\Sigma}_{0|0})$,
    and process noise covariance $\mathbf{Q}$}

    \KwOut{Smoothed trajectory
    $\{\mathbf{x}_{t|T},\boldsymbol{\Sigma}_{t|T}\}_{t=1}^{T}$}

    \CommentSty{// Forward EKF pass} \\

    \For{$t=1,\ldots,T$}{

        \CommentSty{// Motion model} \\

        $\mathbf{x}_{t|t-1}\gets\mathbf{x}_{t-1|t-1}$ \\
        
        $\boldsymbol{\Sigma}_{t|t-1}
        \gets\boldsymbol{\Sigma}_{t-1|t-1}+\mathbf{Q}$ \\

        \CommentSty{// Observations from image} \\

        $\{
            \mathbf{z}_{t,i}^{\mathrm{kp}},
            \mathbf{R}_{t,i}^{\mathrm{kp}}
        \}_{i=1}^{7}
        \gets
        \text{DecodeKeypoints}(\mathbb{H}_t)$ \\

        $\{
            \mathbf{z}_{t,j}^{\mathrm{edge}},
            \mathbf{R}_{t,j}^{\mathrm{edge}}
        \}_{j=1}^{2}
        \gets
        \text{DecodeEdges}(\mathbb{E}_t)$ \\

        $\{
            \mathbf{z}_{t,i}^{\mathrm{mask\text{-}tip}}
        \}_{i=1}^{n_t^{\mathrm{tip}}},
        \{
            \mathbf{z}_{t,j}^{\mathrm{mask\text{-}edge}}
        \}_{j=1}^{n_t^{\mathrm{edge}}}
        \gets
        \text{DecodeMask}(\mathbb{S}_t)$ \\

        \CommentSty{// Observation model} \\

        $\{
            \widehat{\mathbf{z}}_{t,k}^{\mathrm{kp}}
        \}_{k=1}^{12}
        \gets
        \text{ProjectKeypoints}
        (\mathbf{x}_{t|t-1},\boldsymbol{\theta}_t,\mathbf{T}_b^c,\mathbf{K})$ \\

        $\{
            \widehat{\mathbf{z}}_{t,j}^{\mathrm{edge}}
        \}_{j=1}^{2}
        \gets
        \text{ProjectCylinders}
        (\mathbf{x}_{t|t-1},\boldsymbol{\theta}_t,\mathbf{T}_b^c,\mathbf{K})$ \\

        \CommentSty{// NIS-based correspondence selection} \\

        $\{
            \widetilde{\mathbf{z}}_{t,i}^{\mathrm{kp}}
        \}_{i=1}^{7}
        \gets
        \text{MatchKeypoints}_t
        \left(
            \{\mathbf{z}_{t,i}^{\mathrm{kp}},
              \mathbf{R}_{t,i}^{\mathrm{kp}}\}_{i=1}^{7},
            \{\widehat{\mathbf{z}}_{t,k}^{\mathrm{kp}}\}_{k=1}^{12}
        \right)$ \\
        
        $\{
            \widetilde{\mathbf{z}}_{t,j}^{\mathrm{edge}}
        \}_{j=1}^{2}
        \gets
        \text{MatchEdges}_t
        \left(
            \{\mathbf{z}_{t,j}^{\mathrm{edge}},
              \mathbf{R}_{t,j}^{\mathrm{edge}}\}_{j=1}^{2},
            \{\widehat{\mathbf{z}}_{t,j}^{\mathrm{edge}}\}_{j=1}^{2}
        \right)$ \\
        
        $\{
            \widetilde{\mathbf{z}}_{t,i}^{\mathrm{mask\text{-}tip}}
        \}_{i=1}^{2}
        \gets
        \text{MatchGrippers}_t
        \left(
            \{\mathbf{z}_{t,i}^{\mathrm{mask\text{-}tip}}\}_{i=1}^{2},
            \{\widehat{\mathbf{z}}_{t,k}^{\mathrm{kp}}\}_{k=1}^{2}
        \right)$ \\
        
        $\{
            \widetilde{\mathbf{z}}_{t,j}^{\mathrm{mask\text{-}edge}}
        \}_{j=1}^{2}
        \gets
        \text{MatchEdges}_t
        \left(
            \{\mathbf{z}_{t,j}^{\mathrm{mask\text{-}edge}}\}_{j=1}^{2},
            \{\widehat{\mathbf{z}}_{t,j}^{\mathrm{edge}}\}_{j=1}^{2}
        \right)$ \\

        \CommentSty{// Measurement information} \\

        \ForEach{$\nu\in\{\mathrm{kp},\mathrm{edge},\mathrm{mask\text{-}tip},\mathrm{mask\text{-}edge}\}$}{
        
            $\boldsymbol{\Omega}_t^\nu\gets\mathbf{0},
            \quad
            \boldsymbol{\zeta}_t^\nu\gets\mathbf{0}$ \\
        
            \ForEach{matched tuple
            $(\mathbf{z},\widetilde{\mathbf{z}},\mathbf{R})$
            of cue type $\nu$}{
        

                $\mathbf{r} \gets \mathbf{z}-\widetilde{\mathbf{z}}$ \\
                $\mathbf{H} \gets \frac{\partial\widetilde{\mathbf{z}}} {\partial\mathbf{x}_{t|t-1}}$ \\ 
                $\boldsymbol{\Omega}_t^\nu 
                \gets 
                \boldsymbol{\Omega}_t^\nu + \mathbf{H}^{\top}\mathbf{R}^{-1}\mathbf{H}$ \\
                $\boldsymbol{\zeta}_t^\nu 
                \gets 
                \boldsymbol{\zeta}_t^\nu + \mathbf{H}^{\top}\mathbf{R}^{-1}\mathbf{r}$ \\
                
            }
        }

        \CommentSty{// Belief update} \\

        $\mathtt{useMask}_t
        \gets
        \mathds{1}
        \left[
            n_t^{\mathrm{tip}}=2
            \text{ and }
            n_t^{\mathrm{edge}}=2
        \right]$\\

       $\boldsymbol{\Lambda}_t
        \gets
        \boldsymbol{\Sigma}_{t|t-1}^{-1}
        +
        \boldsymbol{\Omega}_t^{\mathrm{kp}}
        +
        \boldsymbol{\Omega}_t^{\mathrm{edge}}
        +
        \mathtt{useMask}_t \cdot
        \left(
            \boldsymbol{\Omega}_t^{\mathrm{mask\text{-}tip}}
            +
            \boldsymbol{\Omega}_t^{\mathrm{mask\text{-}edge}}
        \right)$ \\
        
        $\boldsymbol{\zeta}_t
        \gets
        \boldsymbol{\zeta}_t^{\mathrm{kp}}
        +
        \boldsymbol{\zeta}_t^{\mathrm{edge}}
        +
        \mathtt{useMask}_t \cdot
        \left(
            \boldsymbol{\zeta}_t^{\mathrm{mask\text{-}tip}}
            +
            \boldsymbol{\zeta}_t^{\mathrm{mask\text{-}edge}}
        \right)$ \\

        $\mathbf{x}_{t|t}
        \gets
        \mathbf{x}_{t|t-1}
        +
        \boldsymbol{\Lambda}_t^{-1}\boldsymbol{\zeta}_t$ \\
        
        $\boldsymbol{\Sigma}_{t|t}
        \gets
        \boldsymbol{\Lambda}_t^{-1}$ \\
    }
    
    \CommentSty{// Backward RTS pass} \\

    \For{$t=T-1,\ldots,1$}{

        $\mathbf{A}_t
        \gets
        \boldsymbol{\Sigma}_{t|t}
        \boldsymbol{\Sigma}_{t+1|t}^{-1}$ \\

        $\mathbf{x}_{t|T}
        \gets
        \mathbf{x}_{t|t}
        +
        \mathbf{A}_t
        \left(
            \mathbf{x}_{t+1|T}
            -
            \mathbf{x}_{t+1|t}
        \right)$ \\

        $\boldsymbol{\Sigma}_{t|T}
        \gets
        \boldsymbol{\Sigma}_{t|t}
        +
        \mathbf{A}_t
        \left(
            \boldsymbol{\Sigma}_{t+1|T}
            -
            \boldsymbol{\Sigma}_{t+1|t}
        \right)
        \mathbf{A}_t^\top$ \\
    }

    \Return{
        $\{\mathbf{x}_{t|T},
        \boldsymbol{\Sigma}_{t|T}\}_{t=1}^{T}$
    }
\end{algorithm}

\subsubsection{Sequential Multi-Cue Fusion}

At each frame $t$, the predicted keypoint heatmaps $\{\mathbb{H}_{t,i}\}_{i=1}^7$ and shaft edge heatmaps $\{\mathbb{E}_{t,j}\}_{j=1}^2$ are decoded into image- and Hough-space observations, respectively. Their observation covariances are derived from the corresponding heatmap distributions, reflecting the model’s uncertainty in feature detection, particularly when features are occluded or visually ambiguous. Furthermore, the gripper tip and shaft edge observations are extracted from the predicted mask $\mathbb{S}_t$ with fixed covariances, providing complementary geometric constraints.
 
\textbf{Keypoint observations.} Each keypoint heatmap $\mathbb{H}_{t,i}$, with $i=1,\ldots,7$, is normalized into an image-space distribution
\begin{equation}
    p_{t,i}^{\mathrm{kp}}(\mathbf{u})
    \propto
    \exp\!\left(
        \lambda\mathbb{H}_{t,i}(\mathbf{u})
    \right),
\end{equation}
Inspired by~\cite{park2024onlinesupervised}, the observation noise covariance is constructed from the covariance of this distribution
\begin{equation}
    \mathbf{R}_{t,i}^{\mathrm{kp}}
    =
    s_i
    \sum_{\mathbf{u}}
    p_{t,i}^{\mathrm{kp}}(\mathbf{u})
    \left(\mathbf{u}-\mathbf{z}_{t,i}^{\mathrm{kp}}\right)
    \left(\mathbf{u}-\mathbf{z}_{t,i}^{\mathrm{kp}}\right)^\top
    +\beta_i\mathbf{I},
    \label{eqn:keypoint_observation}
\end{equation}
where the keypoint observation $\mathbf{z}_{t,i}^{\mathrm{kp}} \in \mathbb{R}^2$ is decoded from the expectation, $s_i > 0$ scales the heatmap-derived covariance, and $\beta_i > 0$ accounts for feature localization errors not captured by the heatmap spread.

\textbf{Shaft edge observations.} Similarly, each shaft edge heatmap $\mathbb{E}_{t,j}$, with $j=1,2$, is normalized into a distribution $p_{t,j}^{\mathrm{edge}}(\boldsymbol{\eta})$ over the discretized Hough space and decoded as a line observation $\mathbf{z}_{t,j}^{\mathrm{edge}} \in \mathbb{R}^2$. To account for the ambiguity of line representations, the coordinates are locally unwrapped around $\mathbf{z}_{t,j}^{\mathrm{edge}}$ before computing its covariance
\begin{equation}
    \mathbf{R}_{t,j}^{\mathrm{edge}}
    =
    \mathbf{D}_{\mathrm{edge}}
    \left[
        \sum_{\boldsymbol{\eta}}
        p_{t,j}^{\mathrm{edge}}(\boldsymbol{\eta})
        \boldsymbol{\delta}_{t,j}(\boldsymbol{\eta})
        \boldsymbol{\delta}_{t,j}(\boldsymbol{\eta})^\top
    \right]
    \mathbf{D}_{\mathrm{edge}},
    \label{eqn:shaft_observation}
\end{equation}
where $\boldsymbol{\delta}_{t,j}(\boldsymbol{\eta})$ is the unwrapped difference between $\boldsymbol{\eta}$ and $\mathbf{z}_{t,j}^{\mathrm{edge}}$, and the positive diagonal matrix $\mathbf{D}_{\mathrm{edge}}$ independently scales the orientation and offset uncertainties.

\textbf{Mask-derived observations.} After thresholding and morphological post-processing, the predicted instrument mask $\mathbb{S}_t$ is passed to two separately trained neural networks with ResNet-18 backbones~\cite{he2016deep}, which respectively predict the gripper tips $\{\mathbf{z}_{t,i}^{\mathrm{mask\text{-}tip}}\}_{i=1}^{2}$ and shaft edges $\{\mathbf{z}_{t,j}^{\mathrm{mask\text{-}edge}}\}_{j=1}^{2}$. These observations capture salient geometry from the instrument segmentation, which have been exploited by prior rendering-based methods~\cite{liang2025differentiablerenderingbasedposeestimation, hu2026realtimerenderingbasedsurgicalinstrument}. Unlike the heatmap-derived observations, fixed observation noise covariances $\mathbf{R}^{\mathrm{mask\text{-}tip}}, \mathbf{R}^{\mathrm{mask\text{-}edge}}\in\mathbb{R}^{2\times2}$ are used for the mask-derived cues. To avoid extracting unreliable geometric cues from erroneous mask predictions, these observations are fused only when both networks recover exactly two distinct features.

\textbf{Fusion.} The 9-DoF state, $\mathbf{x}_{t|t}$, is estimated iteratively via EKF and a summary is shown in Alg \ref{alg:multi_cue_fusion}.
First the state is predicted from $\mathbf{x}_{t-1|t-1}$ to $\mathbf{x}_{t|t-1}$ utilizing (\ref{eqn:motion_model}).
Then the predicted state is updated to $\mathbf{x}_{t|t}$ by comparing the decoded observations against the projected observations based on kinematics and camera pinhole geometry \cite{richter2022robotictooltracking}.
Compatible assignments are selected by minimizing their total normalized innovation squared (NIS) under the predicted belief $(\mathbf{x}_{t|t-1}, \boldsymbol{\Sigma_{t|t-1}})$. 
The observations are assumed conditionally independent given the state, allowing their information contributions to be summed in the EKF update~\cite{thrun2002probabilistic}.
The resulting trajectory is then refined through backward RTS smoothing \cite{rauch1965maximum}.

\subsubsection{Pseudo-Label Generation and Self-Training}

Given the smoothed state trajectory $\{\mathbf{x}_{t|T}\}_{t=1}^{T}$, the confidence of each pose estimate is determined from its covariance-normalized agreement with the detected keypoints and shaft edges. At frame $t$, the covariance-normalized residual distances are
\begin{align}
    d_{t,i}^{\mathrm{kp}}
    &=
    \left\|
        \mathbf{z}_{t,i}^{\mathrm{kp}}
        -
        \widetilde{\mathbf{z}}_{t|T,i}^{\mathrm{kp}}
    \right\|_{
        \left(\mathbf{R}_{t,i}^{\mathrm{kp}}\right)^{-1}
    },
    \qquad i=1,\ldots,7,
    \\
    d_{t,j}^{\mathrm{edge}}
    &=
    \left\|
        \mathbf{z}_{t,j}^{\mathrm{edge}}
        -
        \widetilde{\mathbf{z}}_{t|T,j}^{\mathrm{edge}}
    \right\|_{
        \left(\mathbf{R}_{t,j}^{\mathrm{edge}}\right)^{-1}
    },
    \qquad j=1,2,
\end{align}
where $\{\widetilde{\mathbf{z}}_{t|T,i}^{\mathrm{kp}}\}_{i=1}^7$ and $\{\widetilde{\mathbf{z}}_{t|T,j}^{\mathrm{edge}}\}_{j=1}^2$ denote the matched projections obtained from $\mathbf{x}_{t|T}$ following the projection and correspondence procedure similar to Alg.~\ref{alg:multi_cue_fusion}. Covariance normalization penalizes disagreement with confident detections while remaining tolerant to discrepancies from uncertain detections due to occlusion or partial visibility. Let $d_t^{\mathrm{kp}}, d_t^{\mathrm{edge}}$ denote the averaged distances over the visible keypoints and the two shaft edges. The resulting frame confidence is
\begin{equation}
    c_t
    =
    \exp\left[
        -\left(
            \frac{d_t^{\mathrm{kp}}}
                 {\sigma_{\mathrm{kp}}}
        \right)^2
        -\left(
            \frac{d_t^{\mathrm{edge}}}
                 {\sigma_{\mathrm{edge}}}
        \right)^2
    \right].
    \label{eqn:frame_confidence}
\end{equation}
where $\sigma_{\mathrm{kp}}$ and $\sigma_{\mathrm{edge}}$ scale the respective distances. For samples with insufficient valid keypoint correspondences or confidence below a specified threshold $\tau>0$, we set $c_t=0$, so that they do not contribute to self-training.

For each frame, $\mathbf{x}_{t|T}$ is projected to generate a target mask $\mathbb{S}_t^\ast$, keypoint heatmaps $\{\mathbb{H}^\ast_{t,i}\}_{i=1}^{7}$, and edge heatmaps $\{\mathbb{E}^\ast_{t,j}\}_{j=1}^{2}$. Let $\mathcal{D}_r$ denote the resulting real-world dataset with pseudo-labels. 
Following (\ref{eqn:syn_loss}), let $\mathcal{L}_{\mathrm{train},t}$ denote the per-sample training objective evaluated on frame $t$. For the real samples $\mathcal{B}_r\subset\mathcal{D}_r$ in each self-training batch, the normalized confidence-weighted loss is
\begin{equation}
    \mathcal{L}_{\mathrm{real}}
    =
    \frac{1}{|\mathcal{B}_r|}
    \sum_{t\in\mathcal{B}_r}
    \frac{c_t}{\overline{c}}\,
    \mathcal{L}_{\mathrm{train},t},
    \qquad
    \overline{c}
    =
    \frac{1}{|\mathcal{D}_r|}
    \sum_{n\in\mathcal{D}_r} c_n,
    \label{eqn:self_training_loss}
\end{equation}
where $\overline{c}$ is the mean confidence over $\mathcal{D}_r$.
During self-training, each optimization batch combines real-world samples with synthetic replay samples to mitigate tracker-induced bias and preserve the semantics learned during pretraining.


\section{Experiments}

\subsection*{Implementation Setup}

The framework is implemented in PyTorch and evaluated on an NVIDIA RTX 4090 GPU. Synthetic pretraining uses 16,000 training images and 4,000 validation images. The model is pretrained for 100 epochs with a batch size of 32 using AdamW, an initial learning rate of $2\times10^{-4}$, a weight decay of $1\times10^{-4}$, and cosine annealing to $2\times10^{-6}$. The checkpoint at epoch 73, which achieves the lowest validation loss, is used for subsequent sim-to-real adaptation. Self-training is performed separately using unlabeled real-world videos from the SurgPose dataset~\cite{wu2025surgposedatasetarticulatedrobotic} and our collected dataset. The model is fine-tuned for 10 epochs with a batch size of 32 and a fixed learning rate of $2\times10^{-6}$. The backbone uses a learning rate multiplier of $0.1$ during both pretraining and self-training. Each self-training batch contains 25\% real-world samples and 75\% synthetic replay samples. 
Keypoints are decoded as the expectations of the normalized keypoint heatmaps, while shaft edges are decoded from the peak responses of the corresponding heatmaps in Hough space.
The EKF and pseudo-label generation hyperparameters are tuned on the SurgPose self-training split and subsequently reused for the collected unlabeled real-world videos.

\subsection*{Datasets}

The SurgPose dataset~\cite{wu2025surgposedatasetarticulatedrobotic} provides RGB videos synchronized with PSM joint states, together with camera calibration, instrument bounding boxes, and 2D semantic keypoint annotations. We use the RGB images, joint states, and calibration data for self-training, while reserving the provided keypoint annotations for evaluation. In the experiments, 12 source videos of 1,000 frames are used, with both manipulators equipped with LNDs. For each source video, the two instruments are tracked separately, yielding 24 trajectories in total. Inputs to the detection models are preprocessed using a fixed-scale $4{:}3$ crop to keep the target instrument in view while suppressing the other instrument in bi-manual scenes. The resulting predictions are transformed back to the original image for tracking and evaluation.

We use 8 trajectories from source videos 0, 1, 6, and 7 for self-training, covering chicken gizzard and pork liver backgrounds. The remaining 16 trajectories are reserved for evaluation. In the evaluation split, source videos 2--5 and 32--33 share the same backgrounds as the self-training split, while videos 30 and 31 introduce unseen chicken thigh and beef backgrounds, respectively. Videos 30--33 also exhibit more rapid instrument motion.
The ground-truth 2D keypoint annotations are used to evaluate both detection and tracking performance.

In addition, we collect five real-world trajectories with a total of 2,370 frames of synchronized endoscopic images and encoder readings for self-training. The images are rectified at a resolution of $640\times 480$. The trajectories contain relatively structured scenes and a limited range of shaft orientations. For evaluation, we separately annotate 226 real-world images selected to cover different backgrounds, a broader range of shaft orientations and joint configurations, and more challenging visual conditions, including stronger distractions and partial occlusions. For each image, all visible keypoints and both shaft edges are manually annotated. This evaluation set tests whether self-training on relatively constrained unlabeled real-world data generalizes to broader instrument configurations and scene conditions.

\begin{table}[t]
\centering
\caption{\textbf{Keypoint Detection Accuracy.}
Pretrained and self-trained detectors are evaluated on 16 SurgPose trajectories, with errors measured at the original 1400 × 986 resolution and PCK reported as percentages. The best results are shown in bold.}
\label{tab:surgpose_detection_accuracy}
\small
\setlength{\tabcolsep}{4pt}
\begin{tabular}{@{}lccc@{}}
\toprule
Metric & Pretrained & Ours & Improvement\\
\midrule
\multicolumn{4}{@{}l}{\textbf{Per-keypoint accuracy (PCK@20)}} \\
\addlinespace[2pt]
\hspace{1em}Gripper 1
    & 91.59 & \textbf{93.70} & $+2.11$ pp \\
\hspace{1em}Gripper 2
    & 94.92 & \textbf{95.99} & $+1.07$ pp \\
\hspace{1em}End effector
    & 94.13 & \textbf{98.12} & $+3.99$ pp \\
\hspace{1em}Pitch
    & 82.76 & \textbf{91.94} & $+9.18$ pp \\
\hspace{1em}Pitch side
    & 94.60 & \textbf{99.90} & $+5.30$ pp \\
\hspace{1em}Roll
    & 71.72 & \textbf{81.98} & $+10.26$ pp \\
\hspace{1em}Roll side
    & 84.38 & \textbf{98.17} & $+13.79$ pp \\
\midrule
\multicolumn{4}{@{}l}{\textbf{Overall performance}} \\
\addlinespace[2pt]
\hspace{1em}Mean error $\downarrow$ (px)
    & 13.41 & \textbf{10.75} & $-2.66$ px \\
\hspace{1em}P95 error $\downarrow$ (px)
    & 29.15 & \textbf{22.60} & $-6.55$ px \\
\hspace{1em}PCK@10 $\uparrow$
    & 53.77 & \textbf{62.18} & $+8.41$ pp \\
\hspace{1em}PCK@20 $\uparrow$
    & 87.25 & \textbf{93.07} & $+5.82$ pp \\
\hspace{1em}PCK@30 $\uparrow$
    & 95.31 & \textbf{97.27} & $+1.96$ pp \\
\bottomrule
\end{tabular}
\vspace{-0.14in}
\end{table}

\begin{table}[t]
\centering
\caption{\textbf{Real-World Feature Detection Accuracy.}
Pretrained and self-trained detectors are evaluated on the annotated real-world data, with mean keypoint errors reported in pixels and edge detection measured by the modified EA score~\cite{liang2025efficientsurgicalroboticinstrument}. The best and second-best results are shown
in bold and underlined, respectively.}
\label{tab:rw_quantitative}
\footnotesize
\setlength{\tabcolsep}{4pt}
\begin{tabular}{@{}lccc@{}}
\toprule
Model
& Gripper (px) $\downarrow$
& All Kpts. (px) $\downarrow$
& Edge $\uparrow$ \\
\midrule
Unified detector~\cite{liang2025efficientsurgicalroboticinstrument}
& 22.24
& --
& 0.9606 \\

Pretrained
& \underline{17.19}
& \underline{12.57}
& \underline{0.9886} \\


Ours
& \textbf{14.90}
& \textbf{11.38}
& \textbf{0.9887} \\

\bottomrule
\end{tabular}
\vspace{-0.14in}
\end{table}

\subsection{Detection Performance}

Table~\ref{tab:surgpose_detection_accuracy} compares the keypoint detection accuracy of the pretrained and self-trained models on the SurgPose evaluation set. For each ground-truth keypoint annotation $\mathbf{z}_{t,i}^{\mathrm{gt}}$, the corresponding detection $\mathbf{z}_{t,i}^{\mathrm{kp}}$ is evaluated using their Euclidean distance in the original $1400\times986$ image coordinates. The two gripper keypoints are treated as an unordered pair and matched to their ground-truth annotations in each frame. The percentage of correct keypoints (PCK)~\cite{yang2012articulated} is used to compare the detection accuracy of each semantic keypoint. Specifically, for the $i$-th semantic keypoint,
\begin{equation}
    \mathrm{PCK}_i@\tau
    =
    \frac{1}{|\mathcal{V}_i|}
    \sum_{t\in\mathcal{V}_i}
    \mathds{1}\!\left[
        \left\|
            \mathbf{z}_{t,i}^{\mathrm{kp}}
            -
            \mathbf{z}_{t,i}^{\mathrm{gt}}
        \right\|_2
        \leq \tau
    \right],
    \label{eqn:pck}
\end{equation}
where $\mathcal{V}_i$ denotes the set of frames in which $\mathbf{z}_{t,i}^{\mathrm{gt}}$ is available, and $\tau$ is the specified pixel error threshold. 
Self-training improves PCK@20 for all seven semantic keypoints, with the largest gains for the pitch, roll, and roll-side keypoints.

For the overall metrics in Table~\ref{tab:surgpose_detection_accuracy}, the keypoint localization errors are pooled over all visible keypoint instances in the evaluation set. The mean and $95$th-percentile errors are reported to characterize the overall error distribution, and PCK with thresholds $\tau=10,20,30$ are computed over the same pooled set. Self-training reduces both the mean and P95 errors while improving PCK across all evaluated thresholds, demonstrating improved overall localization accuracy. 

Table~\ref{tab:rw_quantitative} evaluates feature detection accuracy on the manually annotated real-world dataset. For the shared gripper tip and shaft edge features, the unified feature detection network of \cite{liang2025efficientsurgicalroboticinstrument} is included as a baseline. Keypoint detection is measured by 2D localization error, while edge detection is evaluated using the modified EA score~\cite{liang2025efficientsurgicalroboticinstrument}, which measures both spatial and angular agreement. Despite using a substantially smaller model and requiring no additional edge refinement steps, the pretrained detector outperforms the baseline on the two shared metrics. Self-training on the collected unlabeled real-world videos further improves all reported metrics. As shown in Fig.~\ref{fig:cover}, the self-trained model produces more reliable detections in both distracted and partially occluded scenes.

\subsection{Pseudo-Label Generation}

Table~\ref{tab:filter_ablation} evaluates how additional observation cues and backward RTS smoothing affect pseudo-label quality. This ablation uses the pretrained detector on the self-training split. Starting with keypoints alone, shaft and mask-derived observations are progressively incorporated to assess their contributions to pseudo-label generation.
Adding shaft observations improves performance, while using all three cues achieves the best results across all metrics. RTS estimates also consistently outperform the corresponding EKF estimates for every observation configuration.

\begin{table}[t]
\centering
\caption{\textbf{Ablation of Multi-Cue Fusion.}
Shaft and mask cues are incrementally incorporated into pseudo-label
generation using the pretrained detector on the training split. Mean keypoint
error and PCK@20 are reported for both EKF and RTS. The best and second-best
results are shown in bold and underlined, respectively.}
\label{tab:filter_ablation}
\footnotesize
\setlength{\tabcolsep}{4pt}
\begin{tabular}{@{}ccccccc@{}}
\toprule
\multicolumn{3}{c}{Observations}
& \multicolumn{2}{c}{Mean error $\downarrow$ (px)}
& \multicolumn{2}{c}{PCK@20 $\uparrow$ (\%)} \\
\cmidrule(lr){1-3}
\cmidrule(lr){4-5}
\cmidrule(lr){6-7}
Kpts. & Shaft & Mask & EKF & RTS & EKF & RTS \\
\midrule
$\checkmark$ & --           & --
    & 11.50 & 10.68 & 91.63 & 93.99 \\
$\checkmark$ & $\checkmark$ & --
    & \underline{11.08} & \underline{10.34}
    & \underline{92.91} & \underline{95.78} \\
$\checkmark$ & $\checkmark$ & $\checkmark$
    & \textbf{10.05} & \textbf{9.22}
    & \textbf{94.36} & \textbf{96.58} \\
\bottomrule
\end{tabular}
\vspace{-0.14in}
\end{table}

\begin{figure*}[!t]
    \centering
    \setlength{\tabcolsep}{0.5pt}
    \renewcommand{\arraystretch}{1}

    \begin{tabular}{
        @{}
        >{\centering\arraybackslash}m{0.12\textwidth}
        *{4}{>{\centering\arraybackslash}m{0.215\textwidth}}
        @{}
    }
        &
        \textbf{\footnotesize Chicken Thigh} &
        \textbf{\footnotesize Beef} &
        \textbf{\footnotesize Chicken Gizzard} &
        \textbf{\footnotesize Pork Liver}
        \\[-1pt]

        \makecell{
            \footnotesize\bfseries CMA-ES + \\[-2pt]
            \footnotesize\bfseries KF~\cite{hu2026realtimerenderingbasedsurgicalinstrument}
        }
        &
        \includegraphics[width=\linewidth]{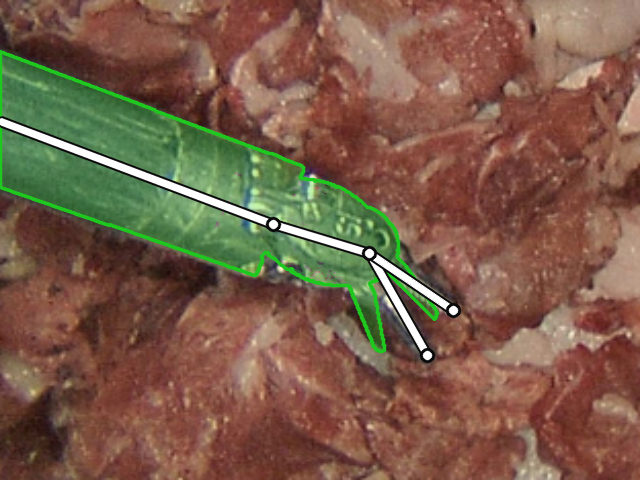}
        &
        \includegraphics[width=\linewidth]{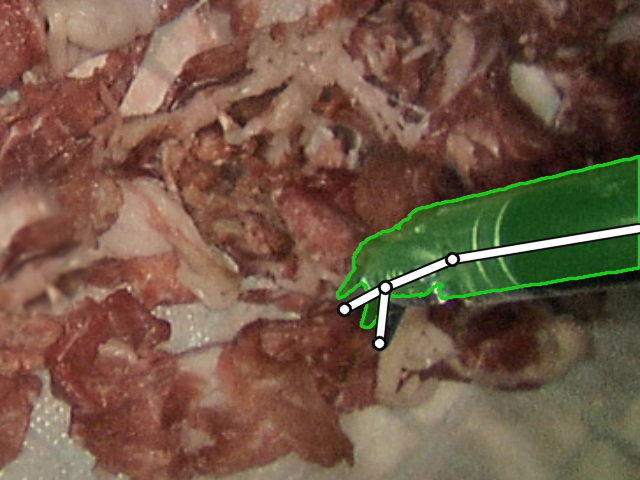}
        &
        \includegraphics[width=\linewidth]{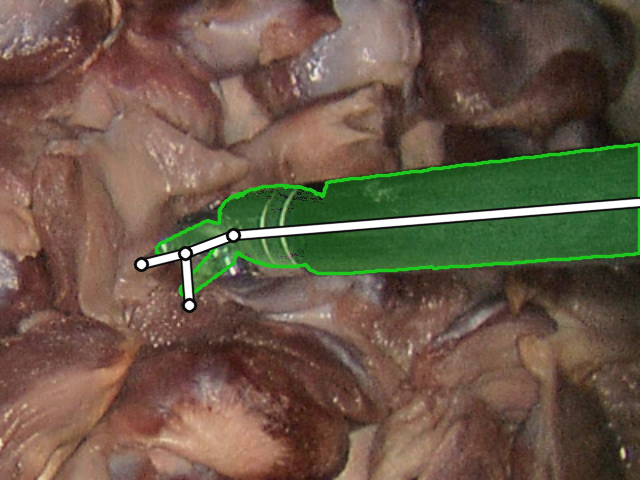}
        &
        \includegraphics[width=\linewidth]{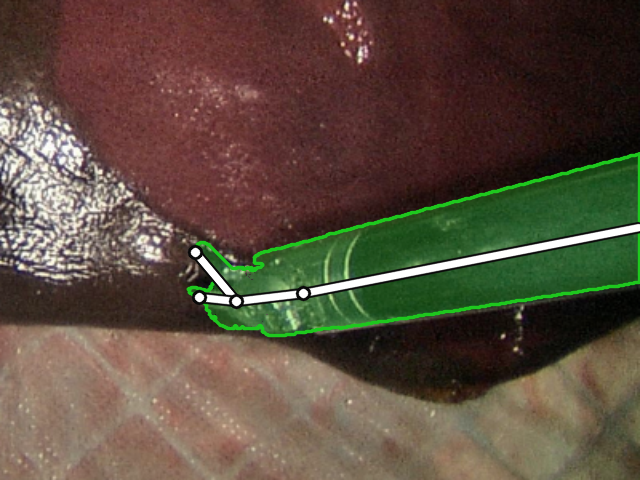}
        \\[-2pt]

        \makecell{
            \footnotesize\bfseries Pretrained\\[-2pt]
            \scriptsize (w/o mask)
        }
        &
        \includegraphics[width=\linewidth]{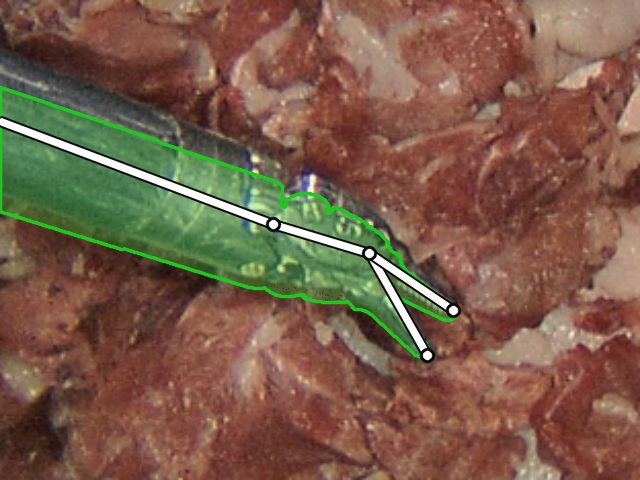}
        &
        \includegraphics[width=\linewidth]{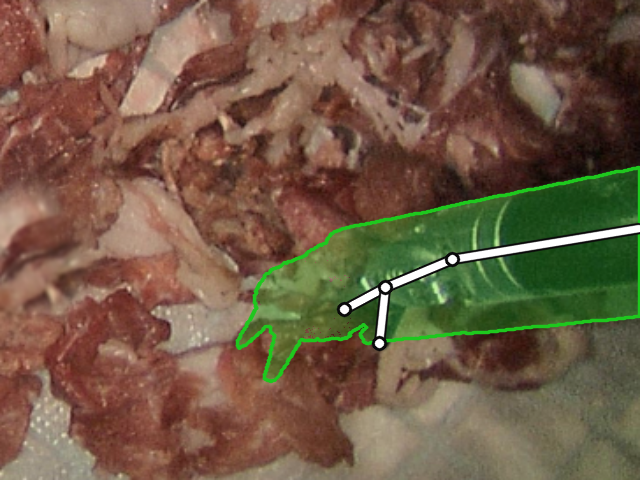}
        &
        \includegraphics[width=\linewidth]{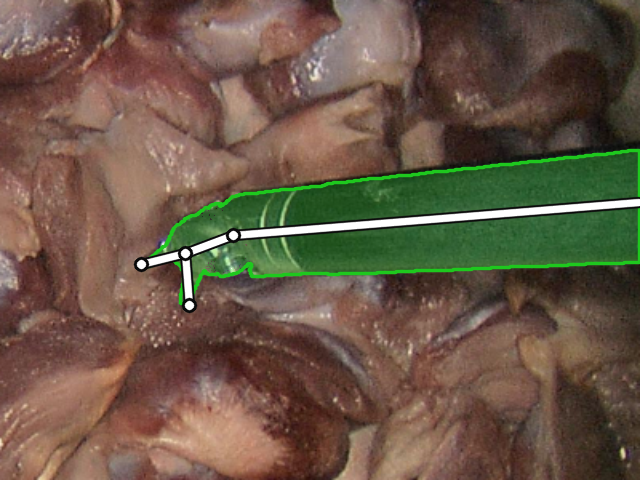}
        &
        \includegraphics[width=\linewidth]{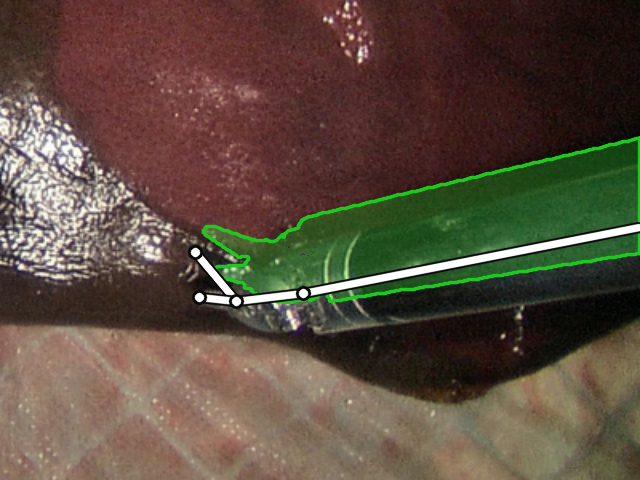}
        \\[-2pt]

        \makecell{
            \footnotesize\bfseries Ours\\[-2pt]
            \scriptsize (w/o mask)
        }
        &
        \includegraphics[width=\linewidth]{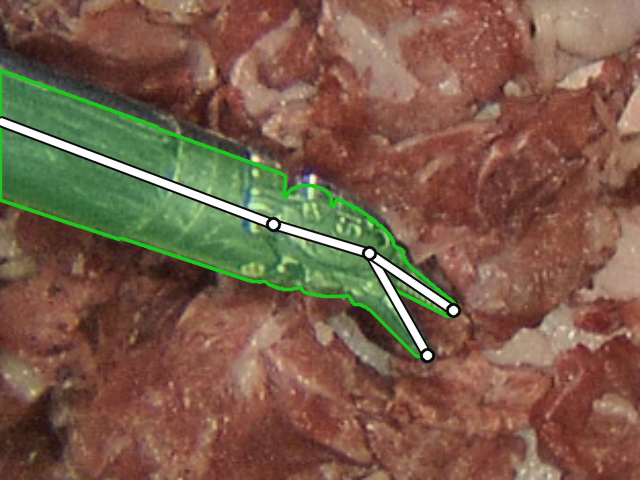}
        &
        \includegraphics[width=\linewidth]{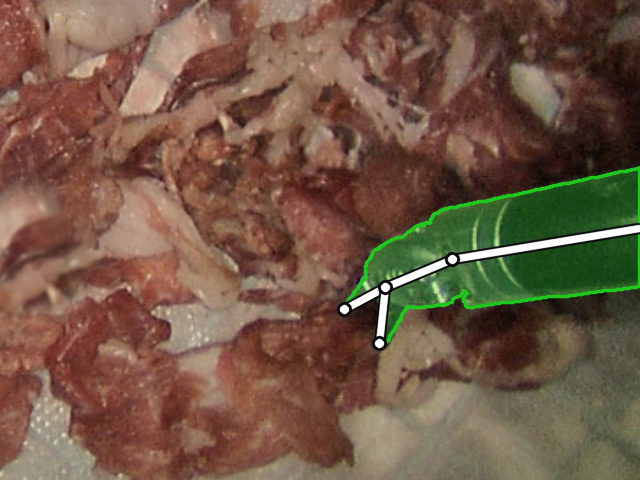}
        &
        \includegraphics[width=\linewidth]{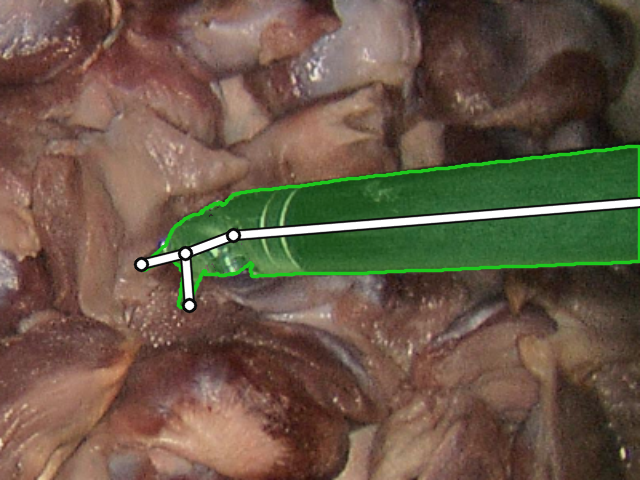}
        &
        \includegraphics[width=\linewidth]{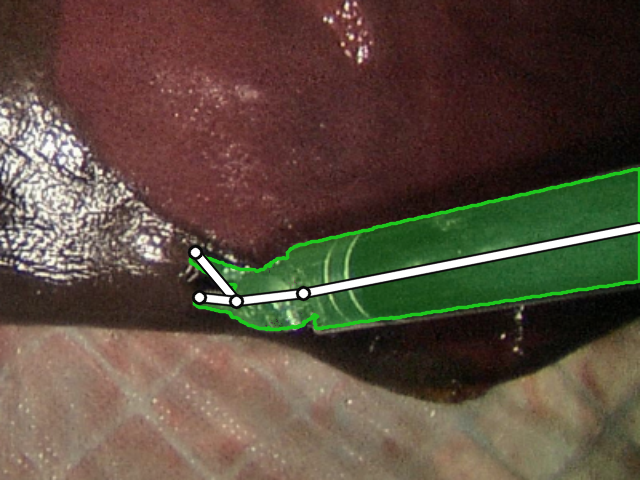}
    \end{tabular}

    \caption{\textbf{Qualitative tracking comparison.} Representative results across different \textit{ex vivo} backgrounds for a rendering-based tracker~\cite{hu2026realtimerenderingbasedsurgicalinstrument} and the proposed EKF tracker with observations from the pretrained and self-trained detectors. 
    The estimated poses are rendered as green silhouettes, while the reference poses are highlighted by white skeletons. 
    The self-trained detector provides more accurate alignment across challenging tissue appearances and unseen backgrounds compared to the rendering-based baseline and the pretrained model.
    }
    \label{fig:qualitative_tracking}
    \vspace{-0.14in}
\end{figure*}

\subsection{Tracking performance.}

Table~\ref{tab:ekf_results} compares the proposed tracker with the rendering-based tracker of \cite{hu2026realtimerenderingbasedsurgicalinstrument}, which uses SurgSAM2~\cite{liu2024surgical} for instrument segmentation and iteratively optimizes instrument state through batched render-and-match evaluation. Tracking accuracy is measured from reprojected keypoints using mean and $95$th-percentile errors and PCK@20, while the reported FPS measures end-to-end throughput for each tracking configuration. For initialization, all methods use the same base-to-camera transform $\mathbf{T}_{b}^{c}$ obtained by rendering-based calibration~\cite{liang2025differentiablerenderingbasedposeestimation, hu2026realtimerenderingbasedsurgicalinstrument}. For the self-trained detector, the scaling factors for the learned keypoint and shaft observation covariances in (\ref{eqn:keypoint_observation}) and (\ref{eqn:shaft_observation}) are reduced to reflect its improved localization accuracy. For trajectories from source videos 30--33 with more rapid instrument motion, the process noise covariance is increased for both detector variants, following the same motivation as the trajectory-specific noise adjustment in \cite{hu2026realtimerenderingbasedsurgicalinstrument}.


The proposed tracker achieves higher accuracy and throughput than the rendering-based baseline, while self-training further improves tracking over the pretrained detector. Mask-derived observations slightly improve tracking with the pretrained detector, but provide no further gain after self-training while substantially reducing throughput. We therefore omit mask-derived observations from the proposed online tracker.
As shown in Fig.~\ref{fig:qualitative_tracking}, the proposed tracker maintains more accurate alignment across challenging tissue appearances and unseen backgrounds, whereas the rendering-based method can degrade under segmentation artifacts and exhibit wrist misalignment due to suboptimal convergence.

\section{Discussion and Conclusions}

\begin{table}[t]
\centering
\caption{\textbf{Tracking Accuracy and Speed.}
Results are reported over the 16 SurgPose evaluation trajectories. The best
and second-best results are shown in bold and underlined, respectively.}
\label{tab:ekf_results}
\footnotesize
\setlength{\tabcolsep}{2.5pt}
\renewcommand{\arraystretch}{1.08}
\begin{tabular}{@{}lcccc@{}}
\toprule
Method
& \shortstack{Mean err. $\downarrow$}
& \shortstack{P95 err. $\downarrow$}
& \shortstack{PCK@20 $\uparrow$}
& FPS $\uparrow$ \\
\midrule
CMA-ES + KF~\cite{hu2026realtimerenderingbasedsurgicalinstrument}
    & 20.53 & 60.47 & 69.17 & 28.78 \\
Pretrained (w/ mask)
    & {9.40} & 22.15 & 93.13 & 81.14 \\
Pretrained (w/o mask)
    & 9.77 & 22.32 & 92.96 & \underline{178.64} \\
Ours (w/ mask)
    & \underline{9.27}
    & \underline{20.60}
    & \underline{94.44}
    & 81.08 \\
Ours (w/o mask)
    & \textbf{9.23}
    & \textbf{20.33}
    & \textbf{94.72}
    & \textbf{179.07} \\
\bottomrule
\end{tabular}
\vspace{-0.14in}
\end{table}

The experimental results highlight two main benefits of the proposed framework. First, self-training consistently improves real-world feature detection, with particularly large gains for keypoints that are less reliably localized by the pretrained model. These improvements also translate to downstream tracking, where the self-trained detector further improves upon the pretrained model and substantially outperforms the rendering-based baseline.
The limited benefit of mask-derived observations after self-training suggests that the improved keypoint and shaft edge predictions already provide sufficient information for accurate tracking.
Importantly, since adaptation operates at the detector level, the resulting feature predictions could also support other feature-based pose estimation and tracking pipelines~\cite{richter2022robotictooltracking,lu2022pose,dambrosia2024robustsurgical,
liang2025efficientsurgicalroboticinstrument}. Second, the pseudo-label ablation shows that incorporating multiple geometric observations and backward RTS smoothing consistently improves the accuracy of the estimated training targets.


Overall, these results demonstrate that sequential multi-cue fusion enables reliable pseudo-label generation from unlabeled real-world videos for improving surgical robotic instrument pose estimation.
Future work will extend the framework to bi-manual scenes and explore robust pose graph optimization to improve pseudo-label generation.

\section*{Acknowledgment}

OpenAI Codex was used to assist with the implementation of the algorithms described in this paper, as well as scripts for visualization and manual data annotation. ChatGPT was used for language polishing and providing feedback on the author-written manuscript text.

\bibliography{IEEEabrv, reference}

\begin{thebibliography}{10}
\providecommand{\url}[1]{#1}
\csname url@samestyle\endcsname
\providecommand{\newblock}{\relax}
\providecommand{\bibinfo}[2]{#2}
\providecommand{\BIBentrySTDinterwordspacing}{\spaceskip=0pt\relax}
\providecommand{\BIBentryALTinterwordstretchfactor}{4}
\providecommand{\BIBentryALTinterwordspacing}{\spaceskip=\fontdimen2\font plus
\BIBentryALTinterwordstretchfactor\fontdimen3\font minus \fontdimen4\font\relax}
\providecommand{\BIBforeignlanguage}[2]{{%
\expandafter\ifx\csname l@#1\endcsname\relax
\typeout{** WARNING: IEEEtran.bst: No hyphenation pattern has been}%
\typeout{** loaded for the language `#1'. Using the pattern for}%
\typeout{** the default language instead.}%
\else
\language=\csname l@#1\endcsname
\fi
#2}}
\providecommand{\BIBdecl}{\relax}
\BIBdecl

\bibitem{molnar2020visual}
C.~Moln{\'a}r, T.~D. Nagy, R.~N. Elek, and T.~Haidegger, ``Visual servoing-based camera control for the da vinci surgical system,'' in \emph{2020 IEEE 18th international symposium on intelligent systems and informatics (SISY)}.\hskip 1em plus 0.5em minus 0.4em\relax IEEE, 2020, pp. 107--112.

\bibitem{kalia2021preclinical}
M.~Kalia, A.~Avinash, N.~Navab, and S.~Salcudean, ``Preclinical evaluation of a markerless, real-time, augmented reality guidance system for robot-assisted radical prostatectomy,'' \emph{International Journal of Computer Assisted Radiology and Surgery}, vol.~16, no.~7, pp. 1181--1188, 2021.

\bibitem{li2020super}
Y.~Li, F.~Richter, J.~Lu, E.~K. Funk, R.~K. Orosco, J.~Zhu, and M.~C. Yip, ``Super: A surgical perception framework for endoscopic tissue manipulation with surgical robotics,'' \emph{IEEE Robotics and Automation Letters}, vol.~5, no.~2, pp. 2294--2301, 2020.

\bibitem{richter2022robotictooltracking}
F.~Richter, J.~Lu, R.~K. Orosco, and M.~C. Yip, ``Robotic tool tracking under partially visible kinematic chain: A unified approach,'' \emph{IEEE Transactions on Robotics}, vol.~38, no.~3, pp. 1653--1670, 2021.

\bibitem{lu2022pose}
J.~Lu, F.~Richter, and M.~C. Yip, ``Pose estimation for robot manipulators via keypoint optimization and sim-to-real transfer,'' \emph{IEEE Robotics and Automation Letters}, vol.~7, no.~2, pp. 4622--4629, 2022.

\bibitem{dambrosia2024robustsurgical}
C.~D’Ambrosia, F.~Richter, Z.-Y. Chiu, N.~Shinde, F.~Liu, H.~I. Christensen, and M.~C. Yip, ``Robust surgical tool tracking with pixel-based probabilities for projected geometric primitives,'' in \emph{2024 IEEE International Conference on Robotics and Automation (ICRA)}.\hskip 1em plus 0.5em minus 0.4em\relax IEEE, 2024, pp. 15\,455--15\,462.

\bibitem{liang2025efficientsurgicalroboticinstrument}
\BIBentryALTinterwordspacing
Z.~Liang, K.~Miyata, X.~Liang, F.~Richter, and M.~C. Yip, ``Efficient surgical robotic instrument pose reconstruction in real world conditions using unified feature detection,'' 2025. [Online]. Available: \url{https://arxiv.org/abs/2510.03532}
\BIBentrySTDinterwordspacing

\bibitem{liang2025differentiablerenderingbasedposeestimation}
Z.~Liang, Z.-Y. Chiu, F.~Richter, and M.~C. Yip, ``Differentiable rendering-based pose estimation for surgical robotic instruments,'' in \emph{2025 IEEE/RSJ International Conference on Intelligent Robots and Systems (IROS)}.\hskip 1em plus 0.5em minus 0.4em\relax IEEE, 2025, pp. 20\,898--20\,905.

\bibitem{chen2025surgipose}
J.-T. Chen, X.~Chen, J.~W. Kim, P.~M. Scheikl, R.~J. Cha, and A.~Krieger, ``Surgipose: Estimating surgical tool kinematics from monocular video for surgical robot learning,'' in \emph{2025 IEEE/RSJ International Conference on Intelligent Robots and Systems (IROS)}.\hskip 1em plus 0.5em minus 0.4em\relax IEEE, 2025, pp. 20\,912--20\,919.

\bibitem{hu2026realtimerenderingbasedsurgicalinstrument}
\BIBentryALTinterwordspacing
H.~Hu, Z.~Liang, F.~Richter, and M.~C. Yip, ``Real-time rendering-based surgical instrument tracking via evolutionary optimization,'' 2026. [Online]. Available: \url{https://arxiv.org/abs/2603.11404}
\BIBentrySTDinterwordspacing

\bibitem{wu2026surgcalib}
Z.~Wu, S.~Yang, Y.~C. Lee, E.~Prisman, Y.~Jin, and S.~E. Salcudean, ``Surgcalib: Gaussian splatting-based hand-eye calibration for robot-assisted minimally invasive surgery,'' \emph{arXiv preprint arXiv:2603.08983}, 2026.

\bibitem{kalman1961new}
R.~E. Kalman and R.~S. Bucy, ``New results in linear filtering and prediction theory,'' 1961.

\bibitem{rauch1965maximum}
H.~E. Rauch, F.~Tung, and C.~T. Striebel, ``Maximum likelihood estimates of linear dynamic systems,'' \emph{AIAA journal}, vol.~3, no.~8, pp. 1445--1450, 1965.

\bibitem{ozguner2020camera}
O.~{\"O}zg{\"u}ner, T.~Shkurti, S.~Huang, R.~Hao, R.~C. Jackson, W.~S. Newman, and M.~C. {\c{C}}avu{\c{s}}o{\u{g}}lu, ``Camera-robot calibration for the da vinci robotic surgery system,'' \emph{IEEE Transactions on Automation Science and Engineering}, vol.~17, no.~4, pp. 2154--2161, 2020.

\bibitem{hwang2020applying}
M.~Hwang, D.~Seita, B.~Thananjeyan, J.~Ichnowski, S.~Paradis, D.~Fer, T.~Low, and K.~Goldberg, ``Applying depth-sensing to automated surgical manipulation with a da vinci robot,'' in \emph{2020 international symposium on medical robotics (ISMR)}.\hskip 1em plus 0.5em minus 0.4em\relax IEEE, 2020, pp. 22--29.

\bibitem{hwang2020efficiently}
M.~Hwang, B.~Thananjeyan, S.~Paradis, D.~Seita, J.~Ichnowski, D.~Fer, T.~Low, and K.~Goldberg, ``Efficiently calibrating cable-driven surgical robots with rgbd fiducial sensing and recurrent neural networks,'' \emph{IEEE Robotics and Automation Letters}, vol.~5, no.~4, pp. 5937--5944, 2020.

\bibitem{wu2021closed}
B.~Wu, L.~Wang, X.~Liu, L.~Wang, and K.~Xu, ``Closed-loop pose control and automated suturing of continuum surgical manipulators with customized wrist markers under stereo vision,'' \emph{IEEE Robotics and Automation Letters}, vol.~6, no.~4, pp. 7137--7144, 2021.

\bibitem{miyasaka2015measurement}
M.~Miyasaka, J.~Matheson, A.~Lewis, and B.~Hannaford, ``Measurement of the cable-pulley coulomb and viscous friction for a cable-driven surgical robotic system,'' in \emph{2015 IEEE/RSJ international conference on intelligent robots and systems (IROS)}.\hskip 1em plus 0.5em minus 0.4em\relax IEEE, 2015, pp. 804--810.

\bibitem{zhong2020hand}
F.~Zhong, Z.~Wang, W.~Chen, K.~He, Y.~Wang, and Y.-H. Liu, ``Hand-eye calibration of surgical instrument for robotic surgery using interactive manipulation,'' \emph{IEEE Robotics and Automation Letters}, vol.~5, no.~2, pp. 1540--1547, 2020.

\bibitem{ye2016real}
M.~Ye, L.~Zhang, S.~Giannarou, and G.-Z. Yang, ``Real-time 3d tracking of articulated tools for robotic surgery,'' in \emph{International conference on medical image computing and computer-assisted intervention}.\hskip 1em plus 0.5em minus 0.4em\relax Springer, 2016, pp. 386--394.

\bibitem{moccia2020vision}
R.~Moccia, C.~Iacono, B.~Siciliano, and F.~Ficuciello, ``Vision-based dynamic virtual fixtures for tools collision avoidance in robotic surgery,'' \emph{IEEE Robotics and Automation Letters}, vol.~5, no.~2, pp. 1650--1655, 2020.

\bibitem{shinde2024surestep}
N.~U. Shinde, Z.-Y. Chiu, F.~Richter, J.~Lim, Y.~Zhi, S.~Herbert, and M.~C. Yip, ``Surestep: An uncertainty-aware trajectory optimization framework to enhance visual tool tracking for robust surgical automation,'' in \emph{2024 IEEE/RSJ International Conference on Intelligent Robots and Systems (IROS)}.\hskip 1em plus 0.5em minus 0.4em\relax IEEE, 2024, pp. 6953--6960.

\bibitem{cui2026ontheflyhandeyecalibrationda}
\BIBentryALTinterwordspacing
Z.~Cui and F.~R. y~Baena, ``On-the-fly hand-eye calibration for the da vinci surgical robot,'' 2026. [Online]. Available: \url{https://arxiv.org/abs/2601.14871}
\BIBentrySTDinterwordspacing

\bibitem{lu2023markerlesscameratorobotposeestimation}
J.~Lu, F.~Richter, and M.~C. Yip, ``Markerless camera-to-robot pose estimation via self-supervised sim-to-real transfer,'' in \emph{2023 IEEE/CVF Conference on Computer Vision and Pattern Recognition (CVPR)}.\hskip 1em plus 0.5em minus 0.4em\relax IEEE, 2023, pp. 21\,296--21\,306.

\bibitem{lu2024ctrnetxcameratorobotposeestimation}
J.~Lu, Z.~Liang, T.~Xie, F.~Richter, S.~Lin, S.~Liu, and M.~C. Yip, ``Ctrnet-x: Camera-to-robot pose estimation in real-world conditions using a single camera,'' in \emph{2025 IEEE International Conference on Robotics and Automation (ICRA)}.\hskip 1em plus 0.5em minus 0.4em\relax IEEE, 2025, pp. 1914--1920.

\bibitem{lu2022slamsupported}
Z.~Lu, Y.~Zhang, K.~Doherty, O.~Severinsen, E.~Yang, and J.~Leonard, ``Slam-supported self-training for 6d object pose estimation. in 2022 ieee,'' in \emph{RSJ International Conference on Intelligent Robots and Systems (IROS)}, pp. 2833--2840.

\bibitem{park2024onlinesupervised}
T.~H. Park and S.~D’Amico, ``Online supervised training of spaceborne vision during proximity operations using adaptive kalman filtering,'' in \emph{2024 IEEE International Conference on Robotics and Automation (ICRA)}.\hskip 1em plus 0.5em minus 0.4em\relax IEEE, 2024, pp. 11\,744--11\,752.

\bibitem{wu2025surgposedatasetarticulatedrobotic}
Z.~Wu, A.~Schmidt, R.~Moore, H.~Zhou, A.~Banks, P.~Kazanzides, and S.~E. Salcudean, ``Surgpose: a dataset for articulated robotic surgical tool pose estimation and tracking,'' in \emph{2025 IEEE International conference on robotics and automation (ICRA)}.\hskip 1em plus 0.5em minus 0.4em\relax IEEE, 2025, pp. 10\,507--10\,514.

\bibitem{dinov3}
\BIBentryALTinterwordspacing
O.~Siméoni, H.~V. Vo, M.~Seitzer, F.~Baldassarre, M.~Oquab, C.~Jose, V.~Khalidov, M.~Szafraniec, S.~Yi, M.~Ramamonjisoa, F.~Massa, D.~Haziza, L.~Wehrstedt, J.~Wang, T.~Darcet, T.~Moutakanni, L.~Sentana, C.~Roberts, A.~Vedaldi, J.~Tolan, J.~Brandt, C.~Couprie, J.~Mairal, H.~Jégou, P.~Labatut, and P.~Bojanowski, ``Dinov3,'' 2025. [Online]. Available: \url{https://arxiv.org/abs/2508.10104}
\BIBentrySTDinterwordspacing

\bibitem{cheng2021perpixelclassificationneedsemantic}
B.~Cheng, A.~Schwing, and A.~Kirillov, ``Per-pixel classification is not all you need for semantic segmentation,'' \emph{Advances in neural information processing systems}, vol.~34, pp. 17\,864--17\,875, 2021.

\bibitem{zhao2021deep}
K.~Zhao, Q.~Han, C.-B. Zhang, J.~Xu, and M.-M. Cheng, ``Deep hough transform for semantic line detection,'' \emph{IEEE Transactions on Pattern Analysis and Machine Intelligence}, vol.~44, no.~9, pp. 4793--4806, 2021.

\bibitem{wang2019adaptive}
X.~Wang, L.~Bo, and L.~Fuxin, ``Adaptive wing loss for robust face alignment via heatmap regression,'' in \emph{2019 IEEE/CVF International Conference on Computer Vision (ICCV)}.\hskip 1em plus 0.5em minus 0.4em\relax IEEE, 2019, pp. 6970--6980.

\bibitem{kazanzides2014open}
P.~Kazanzides, Z.~Chen, A.~Deguet, G.~S. Fischer, R.~H. Taylor, and S.~P. DiMaio, ``An open-source research kit for the da vinci{\textregistered} surgical system,'' in \emph{2014 IEEE international conference on robotics and automation (ICRA)}.\hskip 1em plus 0.5em minus 0.4em\relax IEEE, 2014, pp. 6434--6439.

\bibitem{he2016deep}
K.~He, X.~Zhang, S.~Ren, and J.~Sun, ``Deep residual learning for image recognition,'' in \emph{Proceedings of the IEEE conference on computer vision and pattern recognition}, 2016, pp. 770--778.

\bibitem{thrun2002probabilistic}
S.~Thrun, ``Probabilistic robotics,'' \emph{Communications of the ACM}, vol.~45, no.~3, pp. 52--57, 2002.

\bibitem{yang2012articulated}
Y.~Yang and D.~Ramanan, ``Articulated human detection with flexible mixtures of parts,'' \emph{IEEE transactions on pattern analysis and machine intelligence}, vol.~35, no.~12, pp. 2878--2890, 2012.

\bibitem{liu2024surgical}
H.~Liu, E.~Zhang, J.~Wu, M.~Hong, and Y.~Jin, ``Surgical sam 2: Real-time segment anything in surgical video by efficient frame pruning,'' \emph{arXiv preprint arXiv:2408.07931}, 2024.

\end{thebibliography}

\end{document}